\documentclass[conference]{IEEEtran}
\IEEEoverridecommandlockouts
\usepackage{cite}
\usepackage{amsmath,amssymb,amsfonts}
\usepackage{graphics}
\usepackage{wrapfig}
\usepackage{graphicx}
\usepackage{textcomp}
\usepackage{lipsum}
\usepackage{subcaption}
\usepackage[pagebackref=true,breaklinks=true,letterpaper=true,colorlinks,bookmarks=false]{hyperref}
\usepackage{amsmath}
\usepackage{algorithm}
\usepackage{comment}
\usepackage{algpseudocode}
\usepackage{xcolor}
\def\BibTeX{{\rm B\kern-.05em{\sc i\kern-.025em b}\kern-.08em
    T\kern-.1667em\lower.7ex\hbox{E}\kern-.125emX}}
\begin{document}
\author{\IEEEauthorblockN{Manish Bhattarai}
\IEEEauthorblockA{\textit{University of New Mexico}\\
\textit{Los Alamos National Laboratory}\\
Los Alamos, NM, USA\\
ceodspspectrum@lanl.gov}
\and
\IEEEauthorblockN{ Manel Mart\'inez-Ram\'on}
\IEEEauthorblockA{\textit{University of New Mexico} }\\
Albuquerque,NM\\
manel@unm.edu}

\title{A deep Q-Learning based  Path Planning and Navigation System for Firefighting Environments}

\maketitle
\begin{abstract}
    Live fire creates a dynamic, rapidly changing environment that presents a worthy challenge for deep learning and artificial intelligence methodologies to assist firefighters with scene comprehension in maintaining their situational awareness, tracking and relay of important features necessary for key decisions as they tackle these catastrophic events. We propose a deep Q-learning based agent who is immune to stress induced disorientation and anxiety and thus able to make clear decisions for navigation based on the observed and stored facts in live fire environments. As a proof of concept, we imitate structural fire in a gaming engine called Unreal Engine which enables the interaction of the agent with the environment. The agent is trained with a deep Q-learning algorithm based on a set of rewards and penalties as per it’s actions on the environment.  We exploit experience replay to accelerate the learning process and augment the learning of the agent with human-derived experiences. The agent trained under this deep Q- learning approach outperforms agents trained through alternative path planning systems and demonstrates this methodology as a promising foundation on which to build a path planning navigation assistant capable of safely guiding fire fighters through live fire environments.
\end{abstract}




 

\section{Introduction}


Near-zero visibility, unknown hallways, deadly heat and flame, and people in dire need. These are the challenges firefighters face with every structure fire they respond to. Firefighters endure both extreme external conditions and the internal hazards of stress, panic, and disorientation as part of their daily job. Their central weapon against both internal and external hazards is their training on maintenance of situational awareness or understanding of the activities, and circumstances occurring in one’s immediate vicinity. Maintaining situational awareness is key to a  firefighter’s quick and apt response to an ever-changing environment and is critical to accurate decision-making. Situational awareness can be heavily impacted by both external hazards related to fire, and the corresponding internal stresses experienced by first responders. Loss of situational awareness is one of the main causes in the loss of life of firefighters on scene. Firefighters must make prompt decisions in high-stress environments, constantly assessing the situation, planning their next set of actions, and coordinating with other colleagues, often with an incomplete picture of the situation. Situational awareness is the foundation of further decisions on how to coordinate both rescue operations and fire suppression.  Firefighters on-scene pass their scene interpretations on via portable radio devices to field commanders for further assistance in decision making and the passing along of an inaccurate understanding of current conditions can prove disastrous. The limitation of this decision-making system is well reflected in the annual statistics by the US Fire Administration on the loss of human life\footnote{\href{https://www.usfa.fema.gov/downloads/pdf/publications/ff_fat17.pdf}{https://www.usfa.fema.gov/downloads/pdf/publications/ff\_fat17.pdf}}. Existing fire fighting protocols present an excellent use case for institution of state-of-the-art communication and information technologies to improve search, rescue, and fire suppression activities through improved utilization of the data already being collected on-scene. 
\par
Firefighters often carry various sensors in their equipment, including a thermal camera, gas sensors, and a microphone to assist in maintaining their situational awareness but this data currently is used only in real-time by the firefighter holding the instruments.  Such data holds great potential for improving the capability of the fire teams on the ground if the data produced by these devices could be processed with relevant information extracted and returned to all on scene first responders quickly, efficiently and in real-time in the form of an augmented situational awareness.  
The loss of situational awareness is at the core of disorientation and poor decision making. Advancement in computing technologies, small, cheap, wearable sensors paired with wireless networks combined with advanced computing methodologies such as machine learning (ML) algorithms that can perform all data processing and predicting utilizing mobile computing devices makes it not only possible, but quite feasible to create AI systems that can assist firefighters in understanding their surroundings to combat such disorientation and its consequences. This research presents a theoretical approach that can serve as the backbone upon which such an AI system can be built by demonstrating the power of deep Q-learning in building a path planning and navigation assistant capable of tracking scene changes and offering firefighters alternative routes in dynamically changing fire environments.

\par
AI planning is a paradigm that specializes in design algorithms to solve planning problems. This is accomplished by finding a sequence of actions and addressing the needs and constraints to drive an agent from a specified initial state to a final state satisfying several specified goals. We utilize these paradigms to build a framework that teaches the agent about fire avoidance and deploys a decision process reactive enough to successfully guide the agent through simulated spaces that are as dynamic as those encountered in live fire events. Training in a simulated environment allows us the ability to test a multitude of situations and train the agent for exposure to a vast number of scenarios that would otherwise be impossible in real life. As a result, we get a vastly experienced pilot capable of presenting quick recommendations to a wide variety of situations. The presentation of this technology is meant to serve as the basis upon which to build a navigation assistant in future work. 


\section{Preliminaries}
The work in this paper is based on two distinct fields which are 1) Path planning and navigation and 2) deep reinforcement learning.  
\subsection{Path Planning and Navigation}


A large amount of work focused on path planning and navigation to aid firefighting has been done, but few works address dynamic, continuously changing environments. \cite{su2012path} proposes a mobile robot with various sensors to detect fire sources and use the so called A* search algorithm for rescue. An algorithm based on fire simulation to plan safe trajectories for an unmanned aerial system in a simulator environment is presented in \cite{beachly2018fire}. \cite{jarvis2005robot} shows the efficacy of the covert robotic algorithmic tool for robot navigation in high-risk fire environments. The usage of an ant colony optimization tool to automatically find the safest escape routes in an emergency situation in a simulator environment is shown in \cite{goodwin2015escape} whereas \cite{zhang2020path} formulates the navigation problem as a  "Traveling Salesman" problem and proposes a greedy-algorithm-based route planner to find the safest route to aid firefighters in navigation. \cite{ranaweera2018shortest} proposes a particle swarm optimization for shortest path planning for firefighting robots, whereas \cite{zhang2018intelligent} proposes approximate dynamic programming
to learn the terrain environment and generate the motion policy for optimal path planning for UAV in forest fire scenarios. A methodology for path reconstruction based on the analysis of thermal image sequences is demonstrated in \cite{vadlamani2020novel} which is based on the estimation of camera movement through estimation of the relative orientation with SIFT and Optical flow. \par

Despite the large quantity of work in the literature to aid the firefighters in path planning and navigation, most tend to solve the path planning considering a static environment where a one-time decision is made to guide the agent from source to destination. Such algorithms fail when the environment is dynamic. Furthermore, these algorithms do not allow for the agent to take immediate decisions when encountered with a sudden fire in the path of the chosen navigation path. We propose a deep reinforcement learning-based agent that is capable of taking an instantaneous decision based on learned experiences when subjected to sudden environment changes during navigation. 



\subsection{Deep reinforcement learning}
Reinforcement learning (RL) is a technique that tends to learn an optimal policy by choosing actions based on maximizing the sum of expected rewards. Even though several works exist for path planning in fire environments, no RL based path planning implementations were found for fire scenarios. Outside of the fire scenario, several RL based path planning implementations do exist. \cite{romero2016navigation} demonstrates an RL based navigation of a robot which is provided with a topological map.\cite{li2006q} uses a Q-learning based path planning for an autonomous mobile robot for dynamic obstacle avoidance. 
An RL based complex motion planning for an industrial robot is presented in \cite{meyes2017motion}. 
RL integrated with deep learning has demonstrated phenomenal breakthroughs that are able to surpass human-level intelligence for computer games such as Atari 2600 games \cite{ mnih2013playing} \cite{ mnih2015human} , AlphaGo zero \cite{ silver2017mastering} \cite{tang2017recent} along with various other games. In these frameworks, the AI agent was trained by receiving only the snapshots of the game and game score as inputs. Deep RL (DRL) has also been used for autonomous navigation based on inputs of visual information \cite{surmann2020deep} \cite{kiran2020deep}. \cite{bae2019multi} Proposes a multi-robot path planning algorithm based on deep Q-learning whereas  \cite{lei2018dynamic} demonstrates the autonomous navigation of a robot in a complex environment via path planning based on deep Q-learning(DQL) with SLAM. 
Most of these deep reinforcement learning-based path planning and navigation tasks are based on visual input i.e raw images/depth data which encodes the information about the environment. Based on this information, the navigation agents can establish the relationship between action and the environment. The agent in the DRL system embeds the action-policy map in the modal parameters of the neural nets. \par
\begin{figure*}
    \centering
    \includegraphics[width=1\linewidth]{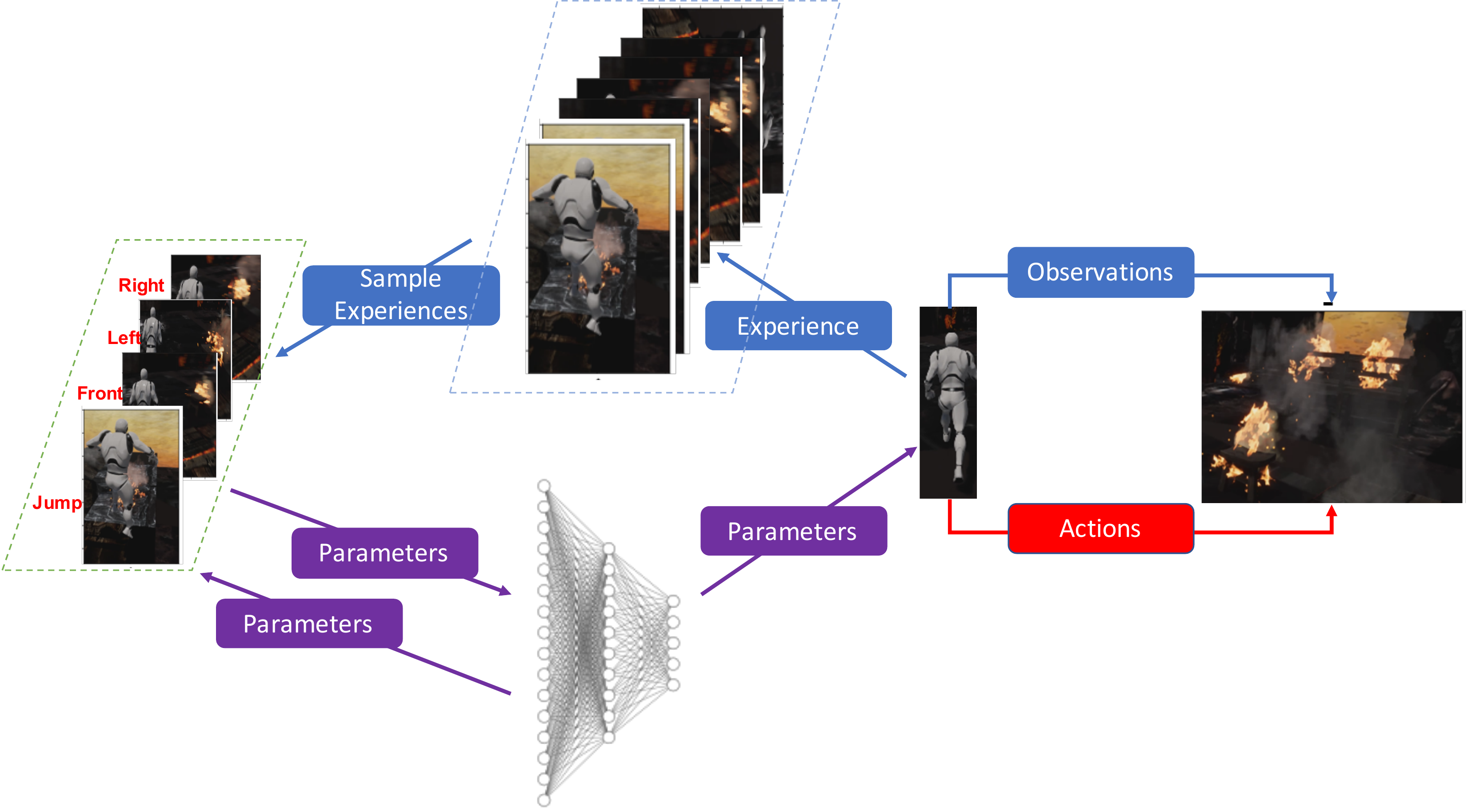}
    \caption{deep Q-Network implementation}
    \label{fig:dqn}
\end{figure*}

Despite the efficacy of the DRL system in navigation, they are based on a learning experience of trial and error where the agent goes through numerous failures before actual success. Considering the hazardous behavior of the fire-environments, training, and evaluation of such DRL systems is very dangerous and practically infeasible. In addition to that, it is very expensive as well as time-consuming. To address these challenges, we developed the training environment for the RL agent in a virtual gaming environment Unreal Engine. The virtual environment depicts the actual firefighting scenario and enables the user to collect a large number of visual observations for action and reaction in various fire environments. The agent can interact with the environment through the actions and can also be trained with various user-defined rewards and goals.  The framework also allows a  plug and play option for the firefighting environment where one can depict a variety of fire scenarios from structural fire to wildfires for training the DRL agent. 
\par
In this paper, we introduce a DRL  approach to train an agent in a simulated fire environment. Taking advantage of simulation, we are able to expose our agent to a vast number of scenarios and dynamics that would be cost as well as safety prohibitive in real life but the results of the training can be applied to real life fire events. The resulting algorithm can be used in conjunction with other deep learning/machine learning approaches to produce a robust navigation assistant that can operate in real-time, effectively guiding fire fighters through a fire scene and aid their decision making by supplementing information gaps and situational awareness lags through the correct interpretation of the scenes they have passed and/or are currently in.


\section{ Problem Description and RL Architecture}
The virtual environment is achieved in a gaming platform Unreal Engine\cite{qiu2016unrealcv} depicting a fire scenario of burning objects and smoke. The gaming engine uses computational fluid dynamics(CFD)\cite{anderson1995computational} based physics models to simulate a real-life dynamic situation where the parameters are a function of time. This gaming environment allows an external interaction where the agent can navigate in the scene via external controls such as a keyboard or head movement in a Virtual reality(VR) device. We take advantage of an interface software AirSim \cite{shah2018airsim} that allows communication to and from the gaming environment to a deep learning framework (Tensorflow \cite{abadi2016tensorflow}). AirSim can grab various parameters from the gaming environment such as RGB feed, Infrared feed, depth, and semantic map information corresponding to the scene and provides the feed to the python block. The python block then processes this information and dictates an agent's movement such as forward, backward, right, left, jump based on the deep reinforcement learning(RL) algorithm, and passes to the AirSim. AirSim further provides these control commands to the Unreal Engine environment which emulates these motions. We deploy a deep Q-learning agent that is trained on a  policy-reward mechanism along with experience replay. For the experience replay, in addition to storing agent self play, we also recorded the user interactions with the environment where they were asked to safely navigate the environment, avoiding the fire and reaching the target in the given scene. With each new start, the user is asked to take different routes with the virtual agent to reach the destination while avoiding fire and the video frames and controls are recorded. During the training of RL agent, the sequence of frames and controls from the experience replay memory are provided to accelerate the training process and make the non-differentiable optimization problem converge in a reasonable time with better accuracy.  The  knowledge gained by the virtual agent on how to successfully navigate the virtual scene can then be transferred to a cyber-human system that can use this knowledge to interpret a real scene and provide step-by-step directions to firefighters to assist them in avoiding fire or other dangerous obstructions. 
The overview of the proposed DQN is shown in Figure \ref{fig:dqn}.
\par
Now, we define our objectives and various parameters associated with the proposed DRL framework. 
\subsection{Objective}
The goal of the proposed deep Q-learning based agent is to reach the destination while safely navigating the fire in a dynamic environment. Safe navigation is defined as avoiding any contact with simulated fire. During the test, the agent needs to be prepared to make instantaneous decisions in instances where fire appears unexpectedly in the chosen navigation path. To achieve the best decisions under such situations, the agents can be subjected to many worst case situations during the training phase. The rewards need to be defined precisely to handle such task-driven learning. 
\subsection{Observations}
The observations for the Q-learning framework are collected through the agent's field of view(FOV) from the virtual simulation environment (i.e Unreal Engine) using the AirSim app. The observations are received by the Python deep learning environment in the form of various feeds which include RGB, infrared, depth, and semantic map frames. Out of these, we are particularly interested in the infrared frames as the CNN framework 
is developed to perform recognition  on thermal imagery. Infrared cameras are the only feed type which can withstand extreme fire and smoke situations and enable a see through smoke. The virtual environment is also able to provide information about the camera position and the agent position, which will be helpful to locate the agent in the given 3D environment. 
\subsection{Actions}
For ease of implementation and demonstration of proof of concept, we have transformed the action space from continuous to discrete space which comprises five primary agent motions. This discretization of the agent space also helps to reduce the model complexity. The five actions are front, back, left, right, and jump. With these motions, the agent can navigate in a structural building containing obstacles like ladders and furniture. The same set of motions also enables the agent to navigate in wildfire scenarios. The agent may take one or a combination of these actions to navigate along the fire scene to reach the destination. 
\subsection{Rewards}
It is very important to define the direction of the goal while training an RL agent. To achieve a task-driven learning objective, it is vital to define the rewards to the navigating agent. The ultimate goal is to find a safe and minimal trajectory length to the navigation target. Unlike the trivial objective of finding the minimal trajectory length, the additional constraint of finding the safe path makes the optimization algorithm more complex. This results in a time-varying decision system whose instantaneous decisions are based on contemporary information of the environment. We introduce a reward and a penalty for familiarizing the agent to the fire environment. The fire have a penalty of -10 and a reward for reaching the goal is 10. 

\subsection{Problem statement}
The RL agent tends to pivot the actions in the direction of maximizing the rewards. The DRL system optimizes the hyperparameters of the Q-neural network to encode the experience of the agent for navigation. The backbone deep network aids the navigation by detection of the objects of interest for navigation. This information is then fed to the Q-network which then chooses the optimal actions to guide the agent. The idea of a DRL system is to provide an end to end learning framework for transforming the pixel information into actions \cite{mnih2015human}. Most of the DRL systems aim to learn the parameters for the neural network to find a transformation from state representations $s$ to policy $\pi(s)$. Also, it is desirable to have an agent that can learn the navigation from a single environment and can generalize the experience to various environments. To achieve that, the aim is to learn a stochastic policy function $\pi$, which can process a representation of current state $s_i$ and target state $s_t$ to produce a probability distribution over action space $\pi(s_i,s_t)$. During the test, the agent samples an action from this distribution until it can achieve the destination. To summarize, the objective function that is used to assess the model performance is given in the form
\begin{equation}
    z = g(x;\theta) = g(x;\beta(\theta);\theta) 
\end{equation}
where $g$ is the navigation problem i.e finding the optimal actions with a DQN of parameters $\theta$ and $\beta$ are the parameters of the navigation agent. $z$ describes the navigation and $J(z)$ measures the optimality of the estimated decisions for navigation. For a given fixed set of neural net parameters $\beta$, the optimizer tends to seek optimal $\theta$ that determines the actions for the agent. To measure the effectiveness of the proposed actions, we use a scoring function $J$ that operates on $\theta$ parameters. 
\begin{figure}[t!] 
\centering
    \includegraphics[width=\linewidth]{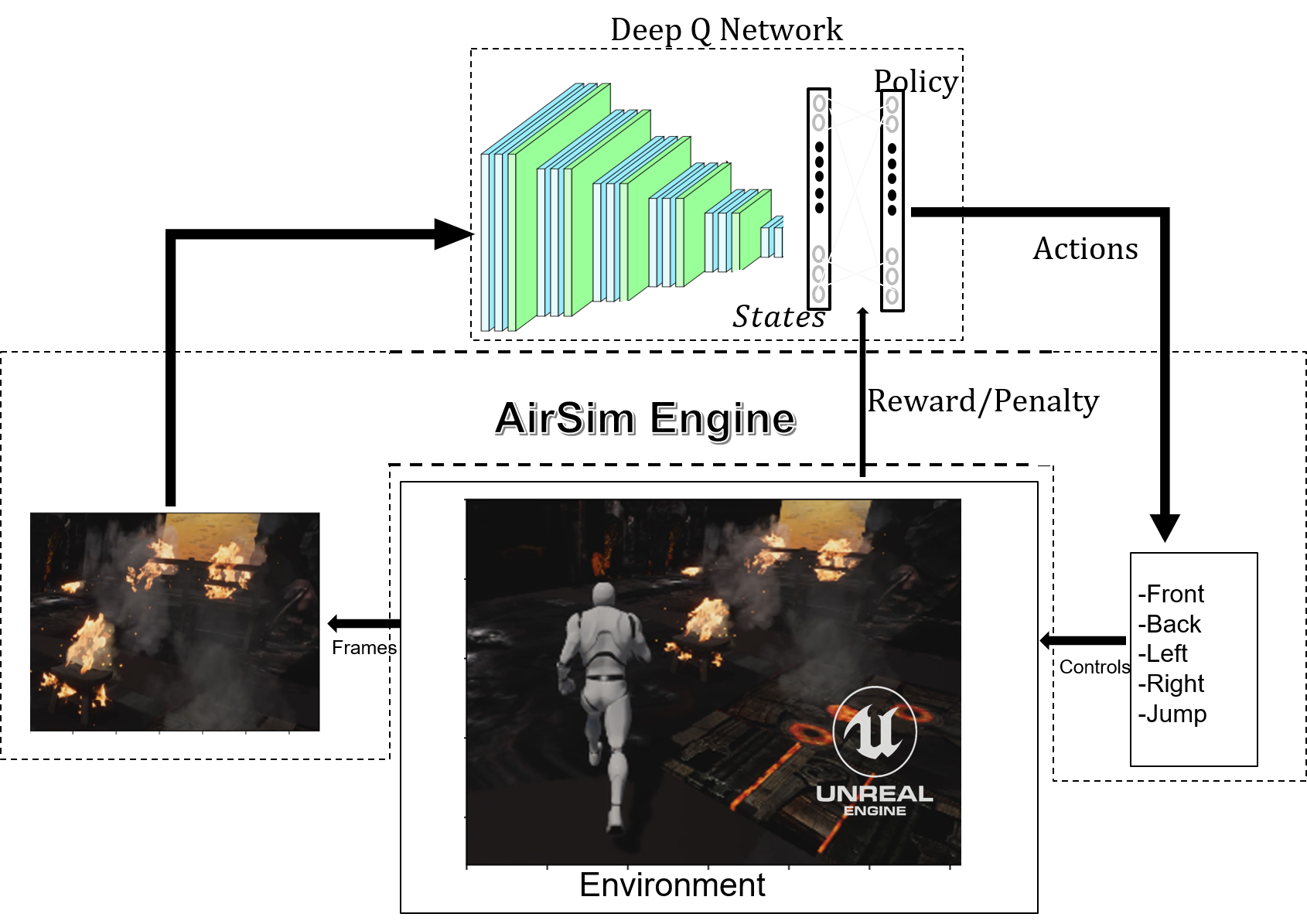}
    \caption{Architecture of Path Planning and Navigation system.  
 }
    \label{fig:sceneunderstanding1}
\end{figure}

\subsection{Model}
The emphasis of this work is to find an optimal policy that can aid a firefighter to navigate in a fire setting via deep reinforcement learning. A deep neural network is trained for a non-linear approximation of the policy function $\pi$, where action $a$ at time $t$ is sampled as :
\begin{equation}
    a \sim \pi(s_i,s_t| \beta) 
\end{equation}
where $\beta$ corresponds to NN model parameters, $s_i$ is the current observation frame, $s_t$ is the target observation to which navigation is to be performed with action sequence $a$. Here, $s_t$ belongs to a discrete set, $\pi$ is a distribution function. The target scene can comprise fire victims which need help for rescue. So, once the DL model estimates the target to be rescued, the RL agent tends to propose navigation paths that successfully rescue the victim. 

\subsection{Q-learning and Deep Q-learning}
We employ a variant of Q-learning called Deep Q-learning (DQL)\cite{mnih2013playing} to train an agent for navigating the fire to reach the destination safely. In this section, we briefly give an overview of the Q-learning and Deep Q-learning algorithms. 
\par
Q-learning learns the action-value function $Q(s, a)$ to quantify the effectiveness of taking an action at a particular state. Q is called the action-value function (or Q-value function ).  In Q-learning, a  lookup table/memory table $Q[s, a]$ is constructed during training to store Q-values for all possible combinations of states $s$ and actions $a$. An action is sampled from the current state, followed by computation of reward R (if any) and then the new state $s’$. From the memory table, the next action $a’$ is determined based on the maximum of $Q(s’, a’)$. 
After this, an action $a$ is performed to seek a reward of $R$. Based on this one-step look ahead, the target $Q(s,a) $ is set to 
\begin{equation}
    target = R(s,a,s') + \gamma max_{a'} Q_k (s',a')
\end{equation}
The update equations are called Bellman equations \cite{bellman1966dynamic} and are performed iteratively with dynamic programming. 
As this update is performed iteratively until convergence, a running average for Q is maintained.


\begin{algorithm}[htp]
    \caption{Deep Q-learning algorithm for path planning agent} \label{alg:q-learning}
    \begin{algorithmic}[1] 
    \State  Initialize replay memory $\mathcal{R}$ to capacity N. 
    \State Initialize the Q-function $Q(s,a)$ for all $s$,$a$ with random weights.
    \For{episode in 1,2,.. M} \Comment{each execution sequence}
    \State Initialize sequence $s_1 = {x_1}$
    \For{t in 1,2..T} \Comment{decision epoch}
    \State With probability $\epsilon$, select a random action, otherwise select 
    $a_t=max_a{Q^*(s,a;\beta)}$ \Comment{\textbf{Exploration vs Exploitation Step}}
    \State Action $a_t$ is performed by agent in the environment and corresponding rewards $r_t$ and scene $x_{t+1}$ is observed. 
    \State Set $s_{t+1} = s_t,a_t,x_{t+1}$ 
    \State Store $s_{t+1}$ in $\mathcal{R}$ . 
    \State Sample a batch of transitions  $e_k = (x_k,a_k,r_k,x_{k+1})$ from $\mathcal{R}$. 
    \If {$x_{t+1}$ is terminal}
    \State $y_k$= $r_k$
    \Else{}
    \State $y_k = r_k + \gamma max_{a'} Q(s_{k+1},a';\beta)$
    \EndIf
    \State Compute loss $(y_k-Q(s_k,a_k;\beta_k))^2$ and then update neural net parameters $\beta$ with gradient descent and back-propagation as per equations 4,5 and 6 .
\EndFor
\State  
\EndFor
     \end{algorithmic}   
\end{algorithm}

However, for solving a real-world problem such as path planning and navigation, where the combinations of states and actions are too large, the memory and the computational requirements for Q is very expensive and intractable in some cases. To address that issue, a deep Q-network (DQN) framework was introduced to approximate Q(s, a) with the aid of neural network parameters. The associated learning algorithm is called Deep Q-learning. Based on this approach. we can approximate the Q-value function with the neural network rather than constructing a memory table for Q-function for state and actions. \par

An RL system needs to know the current state and actions to compute the Q-function. However, for our proposed simulation environment, the internal state information is not available. In one way, the state information can be constructed based on a recognition system that can identify the object of interest in the scene resulting in a discretization of the observation space by assigning pixels discrete values based on their identity. This objective is out of the scope of this paper and will be pursued in the future. For this implementation, we only focus on observing a frame $x_t$ from the emulator, which is a grayscale infrared image.   Based on the action performed in the environment, the agent receives a reward $r_t$, along with a change in the internal state of the environment. Since we have defined a finite reward/penalty corresponding to specific states, the agent might need to go through a series of actions before observing any reward/penalty.

To estimate the Q-function, we consider the sequence of actions and observations for a game play episode. It is given as $s_t=x_1,a_1,x_2,a_2,…..a_{t-1},x_t$. Considering $t$ is a finite time where the game terminates either by reaching the target or getting burnt in a fire, this sequence can be formulated as a markov decision process (MDP).  The goal of the agent is to choose the action that maximizes the sum of future rewards where the reward at time $t$ is given as $R_t=
\sum_{t_1=t}^T \gamma^{t_1-t} r_{t_1}$ for T being episode time. We then use a $Q^*(s,a)$ as optimal action-value function for a given sequence $s$ and action $a$ where
$Q^*(s,a) = max_{\pi} \mathbb{E}[R_t|s_t=s,a_t=a,\pi]$, $\pi$ being the distribution over actions . To estimate this $Q^*(s,a)$, we use the deep neural network (Q-network) of parameters $\beta$ as a non-linear function approximator  in the form $Q(s,a;\beta)$ where it is expected that $Q(s,a;\beta) \approx Q^*(s,a)$.  This network is trained with an objective of minimizing a sequence of loss functions $L_k(\beta_k)$ where,

\begin{equation}
L_k(\beta_k) =\mathbb{E}_{s,a\sim\psi(.)}[(y_k-Q(s,a;\beta_k))^2],
\end{equation}

Where $y_k = E_s’[r+\gamma  max_a’ Q(s’,a’;\beta_{k-1})|s,a]$ is the target for iteration k and $\psi(.)$ is the probability distribution of sequences s and actions a.

The neural net then back-propagates the gradient given as

\begin{equation}
\begin{aligned}
\nabla_{\beta_k} L_i(\beta_k) = \mathbb{E}_{s,a\sim\psi(.)} \Big[\Big(r+\gamma max_{a’} Q(s’,a’;\beta_{k-1}) \\
-Q(s,a;\beta_{k})\Big) \nabla_{\beta_k} Q(s,a;\beta_k)\Big]
\end{aligned}
\end{equation}

The parameters of the neural network are updated as

\begin{equation}
\beta_{k+1} = \beta_{k} - \alpha \nabla_{\beta_k} L_i(\beta_k)
\end{equation}
where $\alpha$ is the learning rate of the neural network.


Furthermore, a technique called experience replay \cite{ mnih2013playing} is used to improve convergence. This occurs through exposing the model to human-controlled navigation and decision making. To implement experience replay, the agent's experience $e_t$ = ($s_t,a_t,r_t,s_{t+1}) $ at each time step $t$  is stored, where $s_t$ is the current state, $a_t$ is the action, $r_t$ is the reward, $s_{t+1}$ is the next state on taking action $a_t$. The experience calculations presented result from the human-controlled navigation training and interaction with the environment. Based on the human interaction, various episodes $\{e_i\}_{i=1}^N$ are then stored in a memory buffer $M$. During the inner loop Q-learning updates, a sample of experiences  are drawn randomly from the memory buffer $M$. The agent then selects and executes an action based on an $\epsilon-$greedy policy. The approach of sampling randomly from experience replay enables the agent to learn more rapidly via improved exposure to reactions to different environmental conditions during an episode of training and allows the model parameters to be updated based on diverse and less correlated state-action data. The algorithm corresponding to the deep Q-learning is presented in Algorithm \ref{alg:q-learning} and implementation methodology is presented in Figure \ref{fig:dqn}.  

\subsection{Network Architecture}
The DQN framework is built on top of a VGG-net-like framework \cite{bhattarai2020deep} as a backbone and is shown in Figure  \ref{fig:sceneunderstanding1}. The backbone framework is used as a feature extractor that produces 4096-d features on a 224x224 infrared/thermal image. The VGG framework is frozen during the training. A stack of 4 history frames are used as state inputs to account for the past sequence of actions of the agent. Then the concatenated feature set comprising $4 \times 4096$ is projected into a 512-d embedding space. This vector is then passed through a fully-connected layer producing 5 policy outputs which give the probability over actions and value output.

\begin{figure*} 
\centering
    \includegraphics[scale=0.6]{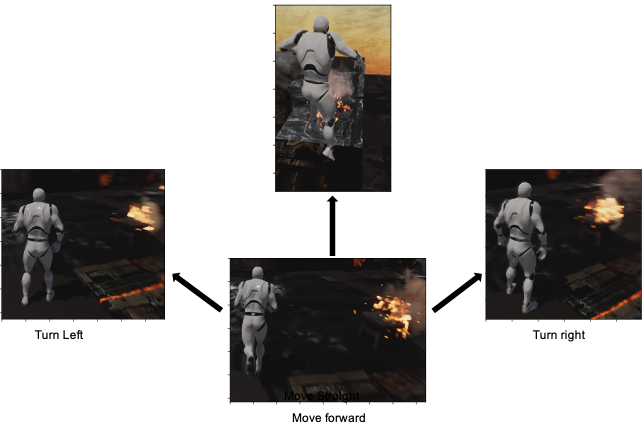}
    \caption{\textbf{ Demonstration Of  Agent Actions In Fire Environment As Dictated By Reinforcement Algorithm.}. Four primary actions demonstrated respectively are turning left, right, jump, and moving forward. This enables the agent to avoid fire and obstacles and safely reach a given destination. 
 }
    \label{fig:sceneunderstanding}
\end{figure*}
\section{Experimental Results}
The implementation of the DQN model was done in Tensorflow \cite{abadi2016tensorflow} on a dual NVIDIA GeForce 1080Ti GPU. The DQN framework is trained with an RMSProp optimizer \cite{tieleman2012lecture} with a learning rate of $10^{-4}$ and batch size of 32. The training was performed with $\epsilon-$ greedy with $\epsilon$ started at 1 and decayed to 0.1 over 5,000 frames. The training afterwards was continued with $\epsilon$ of 0.1. The whole training was performed with as many as 100000 frames with a replay memory of 20,000 frames.  
While training, the gradients are back-propagated from the Q-layer outputs back to lower-level layers while the backbone model is frozen.  The navigation performance was measured by the ability of the agent to reach 100 different targets set in a given fire environment. The target was placed at different locations in the virtual building on different floors where the agent needed to navigate using a combination of all actions. The actions of the agent in successful navigation are shown in Figure \ref{fig:sceneunderstanding}. It is complicated to report the average trajectory length due to the constant changes occurring along the virtual path. In this simulation, it is vital for the agent to avoid fire and the trade off for that is time. Imposing a time constraint as well and weighting the reinforcement model to reward or penalize according to both strictures is a goal for future work.  To prove the efficacy learning deployed by the proposed method, we intensify the situation by increasing fire prevalence. The fire volume per scene was increased from 10\% to 80\% ground coverage of the scene with fires occurring at random across each game. The agent trained using shortest path planning strategy (A*) failed to reach destination when fire coverage reached 30\% while the DQL trained agent was able to consistently navigate across a scene to the destination with a fire coverage up to 76\%. For extreme fire conditions, we carefully increased the rewards and defined additional penalties (distance to fire) to allow a better learning condition for the agent.

\par 
The main goal of the proposed algorithm is to find the least number of combinations of actions that helps the agent to navigate from the current position to the destination while avoiding fire. Due to the dynamic nature of the environment, when we attempted to solve this problem with other path planning techniques including shortest path technique, breadth-first search(BFS) \cite{beamer2012direction}, depth-first search(DFS) \cite{tarjan1972depth}, A*\cite{lavalle2006planning} and random walk\cite{spitzer2013principles}, the probability of the agent reaching the destination was very low (less then 5\%) under the simulation environment.  Since these methods use a single shot decision map to navigate the agent to the destination, the agent was unable to quickly adapt to the continuously varying surroundings. When the agent encountered fire which was not present before the decision, the agent failed to reach the target in most of the cases. In contrast, the agent governed by our proposed method was able to reach the destination with a probability greater then 80\%.  
\par
During the navigation, when the DQL trained agent is unable to find the path to proceed, we design penalties so the agent is  constrained to  either staying in the same position until the path is cleared or retracing its steps backwards to the previous possible path. The authors note this part of the algorithm requires further attention as remaining in one place in a real fire scenario is not realistic. 

\section{Movement planning through deep Q learning for firefighting application}
We have demonstrated the potential deep Q-learning based algorithms hold as a base framework off of which a successful navigation assistant can be built. This methodology can provide an efficient decision-making system for aiding  firefighters who’s decision-making abilities may be impaired due to disorientation, anxiety, and heightened stress levels. This work presents a novel approach to eliminating faulty decisions made under duress through the application of AI planning paradigms.  The paths followed by the firefighters are useful to determine their positions, which is particularly important in search and rescue. 
\par
Existing path planning algorithms can process the information of all paths followed by the firefighters but fail under the constantly changing nature of  the fire ground which can make a previously defined rescue plan unavailable. Also, the presence of smoke and other visual impairments could make difficult the rapid identification of these incidents by a firefighter.  Incidents in the fire ground are hardly predictable by a machine learning system. Machine learning does however, perform well in rapid assessment and production of a decision given the current set of circumstances. In other research outside of the scope of this paper, \cite{bhattarai2020deep,bhattarai2020embedded} have developed a machine learning based methodology that detects and tracks objects of interest  such as doors, ladders, people and fire in the thermal imagery generated by firefighter's thermal cameras. Such information may be valuable to further improve the reinforcement learning algorithm's ability to understand aspects of the environment that may be used in navigation or escape. Future work looks to incorporate a similar object detection work with the path planning work described here to make a robust, navigation assistant that is capable of understanding the surrounding environment outside of fire presence and then recommend best paths to fire fighters. To deploy the agent in a real fire situation, we also aim to first construct the 3D map based on multimodal data(RGB, infrared and depth map) collected from various sensors attached to firefighter's body sensors. Such a map can be imported to the emulator to train the agent in more natural look-a-like environment.   
\par
Assuming that a path has been previously determined by the system by using the information coming from the camera of the rescuer, the rescuer has access to an initial rescue path. The system tells the rescuer to take a direction, which is the present action. The states will be represented by the objects present in the scene. The objects of interest can be represented in a matrix that contains the extracted feature's image. Each detected feature has a different reward. Fire and obstacles have associated penalties, while a clear path has a positive reward. If an obstacle is detected, then the firefighter is told to take a different direction. The new state will be computed for the action taken and a new path will be traced. The recursion repeats until the rescuer has reached the desired position. The plan for this part of the research will include an initial model constructed by simulation. This will be useful to determine the right design for the neural network in terms of stability and convergence speed in different simulated situations. When incorporating information from other paths followed by other fire fighters in a real scene, it is not evident that all obstacles can be determined by their past experiences due to the dynamic nature of a real fire scene. Nevertheless, the parallax estimation obtained from sequences of cameras in motion can be helpful. Parallax data can be used to determine the depth of a given path because it gives the distances between the camera and the key points detected by the SIFT algorithm. This information only needs to be stored and compared with future sequences of the same path. We can consider that an obstacle has been found in a previously clear path if the estimated depth has dramatically changed. In this case. the direction pointed by the camera will be given a low reward instead of a high one.

\section{Conclusion and future directions}

We present a deep Q-learning based agent trained in a virtual environment that is able to make decisions for navigation in an adaptive way in a fire scene. The Unreal engine was used to emulate the fire environment and AirSim was used to communicate data and controls between the virtual environment to the deep learning model. The agent was successfully able to navigate extreme fires based on its acquired knowledge and experience. 

This work serves as the foundation on which to build a deep learning framework that is capable of identifying objects within the environment and incorporating those objects into its decision making process in order to successfully deliver safe, navigable routes to firefighters.  

The learning process is currently slow and needs several hours of training. In the future, we aim to utilize A2C and A3C based reinforcement learning models to train a shared model utilized in parallel by multiple agents with multiple goals simultaneously. we also aim to use the deep learning-based results such as object detection, tracking, and segmentation  to create a more informative situational awareness map of the reconstructed 3d scene. 
\par
The proposed system is intended to be integrated in a geographic and visual environment with data of the floor plan, which will also include scene information about the fire locations, doors, windows, detected firefighters, health condition of the firefighters and other features that are collected from the sensors carried in the fire fighter gear, which will be transmitted over a robust communication system to an incident commander to produce a fully flexed situational awareness system.
\section*{Acknowledgment}
This work was supported by the National Science Foundation (NSF) Smart \& Connected Communities (S\&CC) Early-Concept Grants For Exploratory Research (EAGER) under Grant 1637092. We would like to thank the UNM Center for Advanced Research Computing, supported in part by the National Science Foundation, for providing the high-performance computing, large-scale storage, and visualization resources used in this work. We would also like to thank Sophia Thompson for her valuable suggestions and contributions to the edits of the final drafts.

\bibliographystyle{IEEEtran}
\bibliography{references}

\vspace{12pt}

\end{document}